\documentclass{ieeeaccess}
\usepackage{cite}
\usepackage{amsmath,amssymb,amsfonts}
\usepackage{graphicx}
\usepackage{textcomp}
\usepackage[english]{babel}
\usepackage{amssymb}
\usepackage{mathrsfs,amsmath}
\usepackage[english]{babel}
\usepackage[utf8]{inputenc}
\usepackage[noend]{algpseudocode}
\usepackage{multirow}
\usepackage{graphicx}
\usepackage{url}
\usepackage{soul}
\usepackage{filecontents,lipsum}
\usepackage{dblfloatfix}
\usepackage[utf8]{inputenc}
\usepackage{dtk-logos}

\newtheorem{theorem}{Theorem}[section]

\newtheorem{lemma}[theorem]{Lemma}

\usepackage{placeins}
\makeatletter
\renewcommand{\fnum@figure}{Fig. \thefigure}
\makeatother
\addto\captionsenglish{\renewcommand*{\tablename}{TABLE}}                     
\usepackage{etoolbox}

\def\BibTeX{{\rm B\kern-.05em{\sc i\kern-.025em b}\kern-.08em
    T\kern-.1667em\lower.7ex\hbox{E}\kern-.125emX}}
\begin{document}
\history{Date of publication xxxx 00, 0000, date of current version xxxx 00, 0000.}
\doi{10.1109/ACCESS.2017.DOI}

\title{$\mu$DARTS: Model Uncertainty-Aware Differentiable Architecture Search}
\author{\uppercase{Biswadeep Chakraborty}\authorrefmark{1} and \IEEEmembership{Graduate Student Member, IEEE},
\uppercase{Saibal Mukhopadhyay\authorrefmark{1}},
\IEEEmembership{Fellow, IEEE}}
\address[1]{Department  of  Electrical  and  Computer  Engineering,  Georgia  Institute  of  Technology,   Atlanta,   GA,   30332   USA}

\tfootnote{The research reported here was supported in part by the Defense Advanced Research Projects Agency (DARPA) under contract number HR0011-17-2-0045. The views and conclusions contained herein are those of the authors and should not be interpreted as necessarily representing the official policies or endorsements, either expressed or implied, of DARPA.}

\markboth
{Chakraborty \headeretal: Preparation of Papers for IEEE TRANSACTIONS and JOURNALS}
{Chakraborty \headeretal: Preparation of Papers for IEEE TRANSACTIONS and JOURNALS}

\corresp{Corresponding author: Biswadeep Chakraborty (e-mail: biswadeep@gatech.edu).}
.

\begin{abstract}
We present a \textbf{M}odel \textbf{U}ncertainty-aware \textbf{D}ifferentiable \textbf{AR}chi\textbf{T}ecture \textbf{S}earch ($\mu$DARTS) that optimizes neural networks to simultaneously achieve high accuracy and low uncertainty. We introduce concrete dropout within DARTS cells and include a  Monte-Carlo regularizer within the training loss to optimize the concrete dropout probabilities. A predictive variance term is introduced in the validation loss to enable searching for architecture with minimal model uncertainty. The experiments on CIFAR10, CIFAR100, SVHN, and ImageNet verify the effectiveness of $\mu$DARTS in improving accuracy and reducing uncertainty compared to existing DARTS methods. Moreover, the final architecture obtained from $\mu$DARTS shows higher robustness to noise at the input image and model parameters compared to the architecture obtained from existing DARTS methods.    

\end{abstract}

\titlepgskip=-15pt

\maketitle

\section{Introduction}
Uncertainty estimation of neural networks is a critical challenge for many practical application\cite{gal2016uncertainty, kendall2017uncertainties, kendall2015bayesian}. We can approximate the uncertainty of a neural network by incorporating Monte-Carlo dropout \cite{gal2015dropout}. Gal et al. used concrete dropout as a continuous relaxation of dropout’s discrete masks to improve accuracy and provide better uncertainty calibration ~\cite{gal2017concrete}. For a given training dataset, the model uncertainty of a network depends on its architecture. However, no prior work aims to optimize network architectures to reduce model uncertainty.

We approach this problem by exploring the neural architecture search (NAS)~\cite{bender2019understanding, baker2016designing, bergstra2013making, zoph2016neural,real2019aging}. 
Liu et al. have proposed a differentiable architecture search (DARTS) ~\cite{liu2018darts} that uses continuous relaxation of the architecture representation allowing an efficient architecture search using gradient descent, thereby improving the efficiency of NAS.
The DARTS perform a bi-level optimization problem where an outer loop searches over architectures using a validation loss ($\mathcal{L}_{\text {valid}}$) and the inner loop optimizes parameters (weights) for each architecture using a training loss ($\mathcal{L}_{\text {train}}$). 

We advance the baseline DARTS framework by introducing: (a) concrete dropout ~\cite{gal2017concrete} layers within DARTS cell to enable a well-calibrated uncertainty estimation; (b) a Monte-Carlo dropout based regularizer ~\cite{gal2015dropout} within $\mathcal{L}_{\text {train}}$ to optimize dropout probabilities; and (c) a predictive variance term in $\mathcal{L}_{\text {valid}}$ to search for architecture with minimal model uncertainty. The proposed architecture search is defined as \textbf{M}odel \textbf{U}ncertainty-aware \textbf{D}ifferentiable \textbf{AR}chi\textbf{T}ecture \textbf{S}earch ($\mu$DARTS). This paper makes the following key contributions:

\begin{itemize}
    \item We develop the $\mu$DARTS framework and training process to improve accuracy and simultaneously reduce model uncertainty of neural network.  
    \item We show that the architecture search process in $\mu$DARTS  converges to a flatter minima compared to the standard DARTS method. 
    \item We show that the final architecture obtained via $\mu$DARTS converges to a flatter minima during model training compared to the model obtained using the standard DARTS method. 
    \item We test the final DNN models obtained from architecture search methods on the CIFAR10, CIFAR100, SVHN, and ImageNet datasets. We show that the $\mu$DARTS method improves the accuracy and uncertainty of the final DNN model found using the architecture search.
    \item We also showed that the $\mu$DARTS method has better performance when subjected to input noise and generalizes well when tested with parameter noise
\end{itemize}

This paper aims to find the architecture which not only maximizes the accuracy but also minimizes the predictive uncertainty of the model. We do so using the predictive variance as the regularizer of the bi-level objective function of the DARTS architecture search method. This new architecture search method finds architecture with lower model uncertainty and better generalizability as it induces an implicit regularization on the Hessian of the loss function, leading to more generalizable solutions for the architecture search process.

The rest of the paper is organized as follows: Section II discusses the background and related works to this paper, Section III discusses the theoretical description of the novel Model Uncertainty Aware DARTS methodology proposed in this paper, while Section IV revolves around the implementation details about the architecture search baselines for comparison with the proposed $\mu$DARTS method. Section V deals with the experiments undertaken and the results obtained thereby while finally, Section VI summarizes and discusses the conclusions we arrived at from the experiments.

\section{Background and Related Works}


\subsection{Uncertainty Estimation using Concrete Dropout.} 
Considering the likelihood function of the models to be a multivariate normal distribution given as $p\left(y^{*} | f^{\omega}\left(x^{*}\right)\right)=\mathcal{N}\left(y^{*} ; f^{\omega}\left(x^{*}\right), \Sigma\right)$, we can get an approximate estimate of the expected value of the variational predictive distribution by sampling $T$ sets of weights $\bar{\omega}_{t}$ $(t=1, \ldots, T)$ from the variational dropout distribution\cite{gal2015dropout,gal2017concrete}:
\begin{equation}
    \mathbf{E}_{q_{\theta}\left(y^{*} | x^{*}\right)}\left[y^{*}\right] \approx \frac{1}{T} \sum_{\iota} f^{\tilde{\omega}_{t}}\left(x^{*}\right)
\end{equation}
where $f^{\bar{\omega}_{t}}\left(x^{*}\right)$ denotes a forward pass through the model with the weights which are sampled $\tilde{\omega}_{t},$ thus effectively, performing $T$ forward passes through the network $f$ with dropout and this process is known as Monte Carlo (MC) dropout. 

For multiclass classification tasks with  softmax likelihood $p\left(y^{*} | x^{*}, \omega\right)=\sigma\left(f^{\omega}\left(x^{*}\right)\right),$ we can approximated the variational predictive distribution as:
\begin{equation}
    q_{\theta}\left(y^{*} | x^{*}\right) \approx \frac{1}{T} \sum_{t} \sigma\left(f^{\tilde{\omega}_{t}}\left(x^{*}\right)\right)
\end{equation}


\subsection{Differentiable Architecture Search.} Liu et al. presented DARTS to perform a one-shot neural architectural search. 
The DARTS method ~\cite{liu2018darts} is based on the principle of continuous relaxation of the architecture representation, allowing efficient architectural space search using a gradient descent approach. DARTS formulates the architecture search as a differentiable problem, thus overcoming the scalability challenges faced by other RL-based NAS methods.
The DARTS optimization procedure is defined as a bi-level optimization problem where $\mathcal{L}_{\text {val}}$ is the outer objective  and $\mathcal{L}_{\text {train}}$ is the inner objective as:
\begin{equation}
    \begin{array}{l}
\min _{\alpha} \mathcal{L}_{val}\left(\alpha, w^{*}(\alpha)\right) \\
\text { s.t. } \quad w^{*}(\alpha)={\arg \min }_{w} \mathcal{L}_{\text {train}}(\alpha, w)
\end{array}
\label{eq:DARTS}
\end{equation}

where the validation loss function $\mathcal{L}_{\text {val}}$ determines the architecture parameters $\alpha$(outer variables) and the training loss  $\mathcal{L}_{\text {train}}$ optimizes the network weights $w$ (inner variables). 

The computational graph is learned in one go in the DARTS-based architecture search process. At the end of the search phase, the connections and their associated operations are pruned, keeping the ones that have the highest magnitude of their related architecture weights multipliers. In their paper, Noy et al. \cite{noy2020asap} showed that the harsh pruning in DARTS, which occurs only once at the end of the search
phase, is sub-optimal and that a gradual pruning of connections can improve both the search efficiency and accuracy. On the other hand, Bi et al. \cite{bi2020gold} solve the problem of searching in a complex search space by starting with a complete super-network and gradually pruning out weak operators. Though these methods help in finding efficient architectures, these methods neither estimate nor try to minimize the uncertainty of the model. 


\subsubsection{Improving DARTS Search Space}
Many different works have worked on improving the shortcomings of the DARTS methodologies. For example, in the paper, \cite{chen2019progressive}, the authors pointed out that despite the large gap between the architecture depths in search and evaluation scenarios, the DARTS method report lower accuracy in evaluating the searched architecture or when transferring to another architecture. The authors gradually increased the depth of the searched architectures to address this issue.

Alternatively, in the paper \cite{xu2019pc}, the authors addressed the issue of large memory and computing overheads in jointly training a super-network and searching for an optimal architecture. The authors proposed an approach to sample a small part of the super-network to reduce the redundancy in exploring the network space, thereby performing a more efficient search without comprising the performance.
Again, Dong et al. \cite{dong2019searching} represented the search space as a DAG to reduce the search time of the architecture search process.

This paper uses the DARTS search space and methodology as the baseline. However, our method could very well be extended to perform for any such methodologies mentioned above and search spaces when the aim is to find an architecture in the search space which not only improves the accuracy but also minimizes the uncertainty while making it more robust to noise.

\subsubsection{Improving Robustness and Generalizability of DARTS:} 
Arber et. al. \cite{arber2019understanding} and Chen et.al \cite{chen2020stabilizing} more stabilized neural architecture search methods. SmoothDARTS (SDARTS) use a perturbation-based regularisation to smooth the loss landscape and improve the generalizability.
Empirical results have shown that a generalization performance of the architecture found by DARTS improves with a lower eigenvalue of the Hessian matrix of the validation loss for the architectural parameters ($\nabla_{\alpha}^{2} \mathcal{L}_{\text{val}}^{\text{DARTS}}$).
RobustDARTS \cite{arber2019understanding} has been proposed to improve robustness by (i) computing the  Hessian and stopping DARTS early to limit the eigenvalues (converge to a flatter minima) and (ii) using an L2 regularization term in the training loss ($\mathcal{L}_{\text {train}}$).

\subsection{Neural Architecture Distribution Search:}
Ardywibowo et al. showed that searching for a  distribution of architectures that performs well on a given task allows us to identify standard building blocks among all uncertainty-aware architectures ~\cite{r2020nads}. With this formulation,  the authors optimized a  stochastic out-of-distribution detection (OoD) objective and constructed an ensemble of models  to perform  OoD detection. However, the work concentrates on the detection of out-of-distribution uncertainty by optimizing the Widely Applicable Information Criterion (WAIC), a penalized likelihood score used as the OoD detection criterion. However, they do not discuss the model uncertainty that arises from the generalization error and mainly focus on just the out-of-distribution uncertainty.

\subsection{Contribution of This Work.} The prior works on NAS and DARTS do not focus on minimizing uncertainty. Therefore, the key contribution of this paper is an architecture search method that can simultaneously maximize accuracy and reduce model uncertainty.

\section{Model Uncertainty Aware DARTS}

We propose $\mu$DARTS, a neural architecture search method to find the optimal architecture that simultaneously improves the accuracy and reduces the model uncertainty while also giving a tighter estimate of the model uncertainty using the concrete dropout layers. We formulate the $\mu$DARTS bi-level optimization problem as:
\begin{align}
\min _{\alpha} \mathcal{L}^{\text{DARTS}}_{val}\left(\alpha, w^{*}(\alpha)\right) + \operatorname{Var}_{p(\mathbf{y} | \mathbf{x})}^{model}(\alpha, w^{*}(\alpha)) \nonumber \\
\text { s.t. } \quad w^{*}(\alpha)={\arg \min }_{w} \mathcal{L}^{\text{DARTS}}_{\text {train}}(\alpha, w) + \mathcal{L}_{\mathrm{MC}}(\theta)
\label{eq:muDARTS}
\end{align}

where, $\mathcal{L}^{\text{DARTS}}_{\mathrm{MC}}(\theta)$ is the Monte Carlo dropout loss and $\operatorname{Var}_{p(\mathbf{y} | \mathbf{x})}^{model}(\alpha, w^{*}(\alpha))$ is the predictive variance. As pointed out by Zela et al. \cite{arber2019understanding}, it can be seen that increasing the inner objective regularization helps to control the largest
eigenvalue and more strongly allows it to find solutions with the smaller Hessian spectrum and better generalization properties. Again, we observe the implicit regularization effect on the outer objective, which reduces the overfitting of the architectural parameters. This is further discussed in Appendix A.
Compared to the optimization problem for DARTS (see~\ref{eq:DARTS}), we make the following key updates:  
We know that the validation loss, $\mathcal{L}_{\text{val}}$, is used for the architecture search process, and we want our neural architecture search method to search for the architecture which also has the least predictive uncertainty along with high accuracy. Hence, we add the predictive variance term to the validation loss term and the new validation loss function is given as $\mathcal{L}^{\text{DARTS}}_{val}\left(\alpha, w^{*}(\alpha)\right) + \operatorname{Var}_{p(\mathbf{y} | \mathbf{x})}^{model}(\alpha, w^{*}(\alpha))$, where $w^{*}$ is the optimal set of weights found by optimizing the training loss function in the other bi-level optimization problem. The predictive variance is estimated as the variance of the $T$ Monte Carlo samples on the network as:
\begin{equation}
\operatorname{Var}_{p(\mathbf{y} | \mathbf{x})}^{model}(\mathbf{y})=\sigma_{model}=\sum_{D}\frac{1}{T} \sum_{t=1}^{T}\left(\mathbf{y}_{t}-\bar{\mathbf{y}}\right)^{2}
\label{eq:var}
\end{equation}
where $\left\{\mathbf{y}_{l}\right\}_{t=1}^{T}$ is a set of $T$ sampled outputs for weights instances $\omega^{l} \sim q(\omega ; \Phi)$ and $y=1 / T \sum_{t} y_{t}$.
\par Again, we added a Monte-Carlo loss function in the training loss function since the training loss determines the weights of a particular architecture. To get well-calibrated uncertainty estimates, adapting the dropout probability as a variational parameter to the data at hand is necessary. So, we use the concrete dropout layers and add the Monte Carlo loss to calibrate the dropout probabilities. As shown in \cite{gal2017concrete}, the optimization objective that follows from the variational interpretation can be written as 
    \begin{align}
\mathcal{L}_{\mathrm{MC}}(\theta) &=\frac{1}{N} \mathrm{KL}\left(q_{\theta}(\omega) \| p(\omega)\right) \nonumber \\
\Rightarrow \mathcal{L}^{\text{$\mu$DARTS}}_{\text {train}}(\alpha, w) &= -\frac{1}{M} \sum_{i \in S} \log p\left(\mathbf{y}_{i} |\mathbf{f}^{\omega}\left(\mathbf{x}_{i}\right)\right) + \nonumber\\
&\frac{1}{N} \mathrm{KL}\left(q_{\theta}(\omega) \| p(\omega)\right)
\end{align}
where $\theta$ is the parameters to optimize, $N$ is the number of data points, $S$ is a random set of $M$ data points, $\mathbf{f}^{\omega}\left(\mathbf{x}_{i}\right)$ is the neural network's output on input $\mathbf{x}_{i}$ when evaluated with weight matrices realisation $\omega,$ and $p\left(\mathbf{y}_{i} \mid \mathbf{f}^{\omega}\left(\mathbf{x}_{i}\right)\right)$ is the model's likelihood, e.g. a Gaussian with mean $\mathrm{f}^{\omega}\left(\mathrm{x}_{i}\right) .$ The KL term $\mathrm{KL}\left(q_{\theta}(\omega) \| p(\omega)\right)$ is a ``regularization" term which ensures that the approximate posterior $q_{\theta}(\omega)$ does not deviate too far from the prior distribution $p(\omega) .$

The total derivative of $\mathcal{L}_{\text {val}}$ w.r.t. $\alpha$ evaluated on $\left(\alpha, w^{*}(\alpha)\right)$ would be:
\begin{align}
   \frac{d}{d \alpha} ( \mathcal{L}_{val} + \operatorname{Var}\left(\mathbf{y}^{*}\right) )
   &= \nabla_{\alpha} (\mathcal{L}_{val} + \operatorname{Var}\left(\mathbf{y}^{*}\right)) \nonumber \\ 
   -\nabla_{w} (\mathcal{L}_{val} + \operatorname{Var}\left(\mathbf{y}^{*}\right))&\left(\nabla_{w}^{2} \mathcal{L}_{t r a i n} + \nabla_{w}^{2}\mathcal{L}_{\mathrm{MC}}(\theta)\right)^{-1} \nonumber \\  &\nabla_{\alpha, w}^{2} (\mathcal{L}_{train}+ \mathcal{L}_{\mathrm{MC}}(\theta)) 
\end{align}
where $\nabla_{\alpha}=\frac{\partial}{\partial \alpha}, \nabla_{w}=\frac{\partial}{\partial w}$ and $\nabla_{\alpha, w}^{2}=\frac{\partial^{2}}{\partial \alpha \partial w}.$ 
In general, computing the inverse of the Hessian is not possible considering the high dimensionality of the model parameters $w$. Thus, we use gradient-based iterative algorithms to find the optimal $w^*$. 
However, to avoid repeated training of each architecture which is computationally very expensive, we approximate $w^*(\alpha)$ by updating the current model parameters $w$ using a single gradient descent similar to the approximation step done in ~\cite{liu2018darts}:
\begin{align}
    w^{*}(\alpha) \approx w-\xi \nabla_{w} [\mathcal{L}_{t r a i n}(\alpha, w) + \mathcal{L}_{\mathrm{MC}}(\theta)] \\ \nonumber
    \Rightarrow \frac{\partial w^{*}}{\partial \alpha}(\alpha) = -\xi \nabla_{\alpha, w}^{2} [\mathcal{L}_{t r a i n}(\alpha, w) + \mathcal{L}_{\mathrm{MC}}(\theta)] 
\end{align}
where $\xi$ is the learning rate for the virtual gradient step, DARTS takes with respect to the model weights $w$.
Therefore, we obtain:
\begin{align}
    &\frac{d}{d \alpha}\mathcal({L}_{val} + \operatorname{Var}\left(\mathbf{y}^{*}\right)) =\nabla_{\alpha} \mathcal{L}_{\text {val}}\left(\alpha, w^{*}\right) + \nabla_{\alpha} \operatorname{Var}\left(\mathbf{y}^{*}\right) \nonumber\\ &-\xi \nabla_{w} [\mathcal{L}_{\text {val}}\left(\alpha, w^{*}\right) + \operatorname{Var}\left(\mathbf{y}^{*}\right)] \nabla_{\alpha, w}^{2} (\mathcal{L}_{\text {train}}\left(\alpha, w^{*}\right) + \mathcal{L}_{\mathrm{MC}}(\theta) )
\end{align}
where the inverse Hessian $\nabla_{w}^{2} \mathcal{L}_{t r a i n}^{-1}$ is replaced by the learning rate $\xi$.

The final output using the $\mu$DARTS method gives us the optimal architecture, which has the maximum accuracy and the minimum uncertainty in the architecture search space. However, there are some benefits to including the predictive variance term in the validation loss and the Monte Carlo dropout loss. We hypothesize a two-fold benefit from this modified loss function structure.

Firstly, the predictive variance term acts as a regularizer to the validation loss function. It makes the neural architecture search method more robust and resilient to input or parameter noise. In this paper, we shall prove this hypothesis empirically by computing the largest eigenvalue of the Hessian of the validation loss to indicate the flatness of the loss minima of the architecture search process. Also, analytical proof of this is given in Appendix B, considering a simplified linear model.

Similarly, the Monte Carlo dropout loss added to the training loss function acts as another regularizer. Hence, the final architecture obtained from the neural architecture search method also gives better performance under noise perturbation. Similarly, we also prove this empirically by computing the largest eigenvalue of the Hessian of the training loss to indicate that the architecture converges to a flat minima for the final architecture we got from the neural architecture search process.

\section{Implementation of Baselines and $\mu$DARTS}

\begin{figure}[ht]
\includegraphics[width = 0.9\columnwidth]{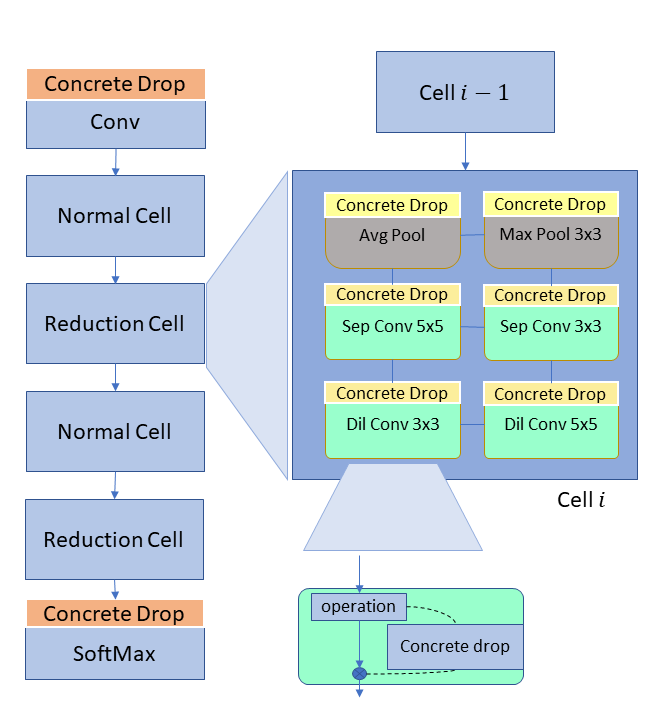}
\centering
\caption{Flowchart of the $\mu$DARTS method}
\centering
\label{fig:flowchart}
\end{figure}

\begin{figure*}
\includegraphics[width =0.9\textwidth]{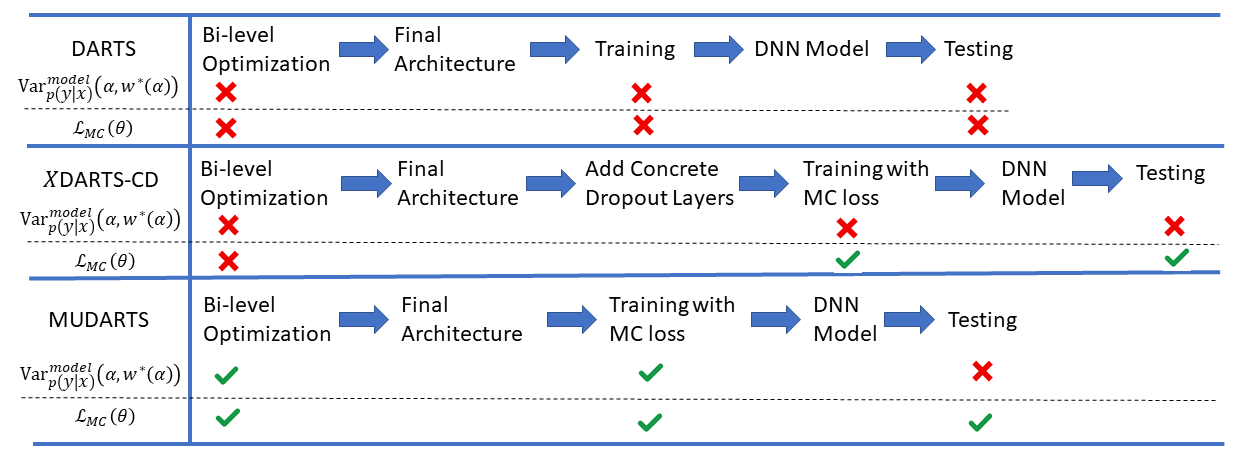}
\centering
\caption{Summary of Baselines and $\mu$DARTS. $X$DARTS-CD refers to the baseline architecture search models viz., P-DARTS-CD, PC-DARTS-CD, RobustDARTS-CD, GOLD-NAS-CD, ASAP-CD described in Section IV}
\centering
\label{fig:flowchart2}
\end{figure*}

\textbf{DARTS:} We implemented the standard DARTS method \cite{liu2018darts} with the search space constituting of the following operations: $\mathcal{O}: 3 \times 3,\quad5 \times 5$ separable convolutions (sep conv), $3 \times 3,\quad 5 \times 5$ dilated separable convolutions (dil conv), $3 \times 3$ max pooling, $3 \times 3$ average pooling, identity, and zero (Fig. \ref{fig:flowchart}). All the operations in the method are of unit stride length, wherever applicable, and paddings are added to the convolved feature maps to preserve their spatial resolution. For the convolution operations, we use the order $\texttt{ReLU/Conc Drop(ReLu)-Conv-BN}$ as done in ~\cite{zoph2016neural, real2019aging}. The output node of the convolutional cells is the depthwise concatenation of all the intermediary nodes except the input nodes. The network is then formed by stacking multiple cells following the same principle used in \cite{liu2018darts}.
We used the implementation of DARTS used in \cite{liu2018darts} for our results and ran it for 50 epochs for each searching and training algorithm.

\textbf{DARTS with Concrete Dropout (DARTS-CD).} 
The final architecture obtained from DARTS cannot be directly used for uncertainty estimation. We modify the optimal architecture from the standard DARTS to enable uncertainty estimation. We include a Concrete dropout \cite{gal2017concrete} layer in all the layers of the final architecture, and the DARTS-CD is obtained by adding a Monte Carlo dropout loss to the training loss function $\mathcal{L}_{\text{train}}$ to calculate the optimal dropout probabilities~\cite{gal2017concrete}. We use the re-trained final architecture to estimate accuracy and uncertainty using multiple MC samples. 

\textbf{RobustDARTS with Concrete Dropout (RDARTS-CD).} We implemented the RobustDARTS (RDARTS) model using the L2 regularization in the training loss, and the early stopping mechanism \cite{arber2019understanding}. The RDARTS generates a final architecture. Like DARTS-CD, concrete dropout layers are included in each of the layers in RDARTS for uncertainty estimation. The modified architecture is re-trained by adding Monte-Carlo dropout loss in the loss function. The modified final architecture from RDARTS calculates the model uncertainty and accuracy.

\textbf{Progressive Differentiable Architecture Search with Concrete Dropout (P-DARTS-CD)}
We also implemented the Progressive Differentiable Architecture Search (P-DARTS) \cite{chen2019progressive}. P-DARTS is an efficient algorithm that gradually increases the depth of the searched architectures while training. P-DARTS solves the problems of heavier computational overheads and weaker search stability using search space approximation and
regularization, respectively. In this paper, we added Concrete Dropout after each layer of the P-DARTS searched architecture to get an uncertainty estimate using the Monte Carlo Dropout methodology. The modified network is trained with added Monte Carlo dropout loss. The modified final architecture from P-DARTS is then used to calculate the accuracy and model uncertainty.

\textbf{PC-DARTS-CD}
Partially-Connected DARTS (PC-DARTS)\cite{xu2019pc} performs a more efficient search without comprising the performance by sampling a small part of the super-network to reduce the redundancy in exploring the network space. Similar to the previously mentioned architectures, we add Concrete Dropout layers to the final architecture from the PC-DARTS method and, using the Monte Carlo Dropout loss in the loss function, get an estimate of the uncertainty of the model.

\textbf{GOLD-NAS-CD} Gradual One-Level Differentiable Neural Architecture Search (GOLD-NAS) introduces a variable resource constraint to one-level optimization so that the weak operators are gradually pruned out from the super-network. Similar to the other methods, we add additional Concrete Dropout layers after each layer in the final architecture to get the uncertainty estimate of the model.

\textbf{ASAP-CD}  Architecture Search, Anneal and Prune (ASAP) \cite{noy2020asap} uses a differentiable search space that allows the annealing of architecture weights while gradually pruning inferior operations. In this way, the search converges to a single output network continuously. Similar to the other methods, we add Concrete Dropout layers after each layer in the final architecture obtained in the paper to get the uncertainty estimate of the model.

$\mathbf{\mu}$\textbf{DARTS.} Fig. \ref{fig:flowchart} shows the overall architecture of the proposed $\mu$DARTS method, including the internal details of a cell. 
We include the following operations in $\mathcal{O}$: $3\times 3, 5 \times 5$ separable convolutions (sep conv), $3 \times 3, 5 \times 5$ dilated separable convolutions (dil conv), $3\times3$ max pooling, $3\times3$ average pooling, identity, and zero. 
The key difference between an $\mu$DARTS cell compared to the standard DARTS cell is that each cell has an option to have a Concrete dropout layer. The Concrete dropout layer, if included, enables computation of the uncertainty values of the model. Wherever applicable, the operations are of unit stride, and also we use padding in the convolved feature maps. $\mu$DARTS also includes a Concrete dropout in the final softmax layer of the model.

The comprehensive details for all the methods described above are shown in Fig. \ref{fig:flowchart2}.

\section{Experimental Results}

\subsection{Training Conditions}

The experiments performed in this paper were performed on a single NVIDIA GeForce GTX 1080 Ti Graphics Card. To get a fair comparison between the models, the experiments performed in this paper keep the training parameters constant for all the models. We trained all the methods for $50$ epochs for the architecture search process with a batch size of $32$ and a learning rate of $0.05$. After getting the final architecture, we train the final model for $50$ epochs with a batch size of $64$ and a learning rate of $0.025$ to get the best results of the model obtained. Table \ref{tab:hyperparams} summarizes the hyperparameters used in the architecture search.  

\begin{table}[]
\centering
\caption{Table showing the hyperparameters of the architecture search process}
\label{tab:hyperparams}
\resizebox{0.4\columnwidth}{!}{%
\begin{tabular}{|c|c|}
\hline
\textbf{Hyperparameters} & \textbf{Values} \\ \hline
Learning Rate & 0.025 \\ \hline
Weight Decay & 0.0243 \\ \hline
Epochs & 50 \\ \hline
Batch Size & 16 \\ \hline
No. of cells & 14 \\ \hline
\end{tabular}%
}
\end{table}

\par The experimental results of the paper are divided into the following subsections:
\begin{itemize}
    \item Comparative analysis of the architecture search methods, including ablation studies, robustness, convergence, and run-time.
    \item Comparative analysis of the final architecture, including flatness of the loss surface and testing errors.   
    \item Comparative analysis of the final DNN models, including accuracy and uncertainty comparisons and tolerance to input and parameter noise.
\end{itemize}

In the rest of this paper, we primarily train the models under the same training conditions mentioned above for uniformity and evaluate them to get the accuracy and uncertainty estimates. For reference, we used the vanilla models (without adding concrete dropout layers), as implemented in the original papers, and compared them with the accuracy and training condition of $\mu$DARTS as implemented in this paper. The results are shown in Appendix C.

\subsection{Analysis of the Architecture Search Methods}

\begin{figure*}
    \centering
    \includegraphics[width=\textwidth]{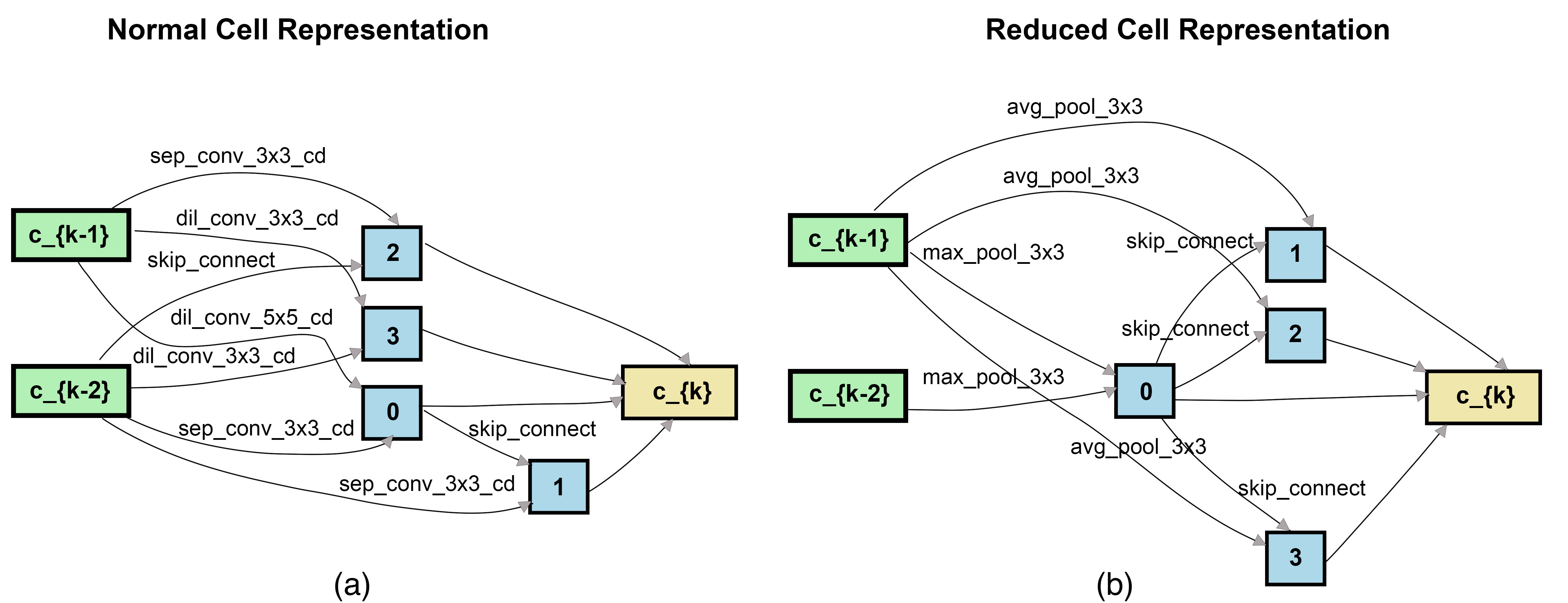}
    \caption{Figure showing the normal and reduction cell representations searched using the $\mu$DARTS algorithm}
    \label{fig:arch}
\end{figure*}

The searched architecture using the $\mu$DARTS architecture search method is shown in Fig. \ref{fig:arch}

\subsubsection{Ablation Studies}

In this section, we perform an ablation study for the two different cases: 1. removing the Monte Carlo loss function $\mathcal{L}_{\mathrm{MC}}(\theta)$ from the training loss $\mathcal{L}^{\text{DARTS}}_{val}$ 2. removing the prediction variance term $\operatorname{Var}_{p(\mathbf{y} | \mathbf{x})}^{model}(\alpha, w^{*}(\alpha))$ from the validation loss term $\mathcal{L}^{\text{DARTS}}_{train}$.

\begin{figure}
    \centering
    \includegraphics[width =\columnwidth]{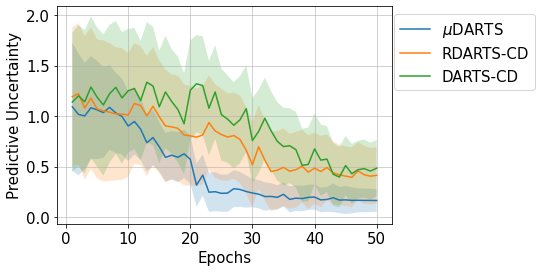}
    \caption{Plot showing the mean and variance for the variation of model uncertainty, measured using the predictive variance for different architecture search methods with increasing epochs  for CIFAR 10 dataset}
    \label{fig:ablation1}
\end{figure}

\textbf{Role of  $\operatorname{Var}_{p(\mathbf{y} | \mathbf{x})}^{model}(\alpha, w^{*}(\alpha))$:}
Including $\mathcal{L}_{\mathrm{MC}}(\theta)$ results in calibrated uncertainty values in the inner loop of the DARTS optimization loop. However, if the predictive variance term is removed, the neural architecture search method is not optimized for the uncertainty values. Removing the predictive variance loss term but not the MC loss is equivalent to the DARTS-CD architecture search method. We simulate this ablation study on the CIFAR 10 dataset and compare the performance with the case where we include the predictive variance term in the training loss function. We plot the variation of the predictive uncertainty with increasing epochs for $\mu$DARTS and DARTS-CD in Fig. \ref{fig:ablation1}.  

\begin{figure}
    \centering
    \includegraphics[width = \columnwidth]{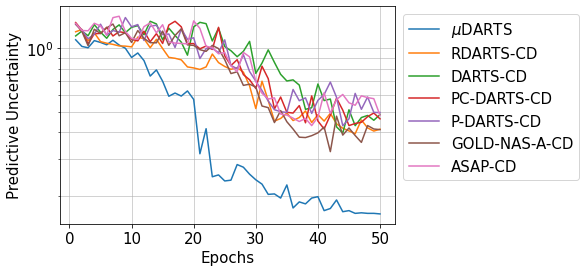}
    \caption{Plot showing the variation of predictive variance for different architecture search methods with increasing epochs  for CIFAR 10 dataset (Note that the y-axis is represented in the logarithmic scale)}
    \label{fig:pred_var}
\end{figure}

In this paper, we search for an architecture that not only maximizes accuracy but simultaneously minimizes the model uncertainty. We accomplish this by using the predictive variance as a regularizer of the loss function as given in Eq. 4.
To verify that adding the predictive variance term helps in searching for architectures with lower predictive uncertainty, we plotted the evolution of the predictive uncertainty of different architecture search results with respect to $\mu$DARTS. The results are shown in Fig. \ref{fig:pred_var}. 
We see that the model uncertainty (measured using the predictive variance) remains almost constant for models like DARTS-CD and RDARTS-CD, which steadily decreases for the $\mu$DARTS architecture search method, which uses the predictive variance as a regularizer of the Cross-Entropy loss function.

\begin{figure}
    \centering
    \includegraphics[width = 0.8\columnwidth]{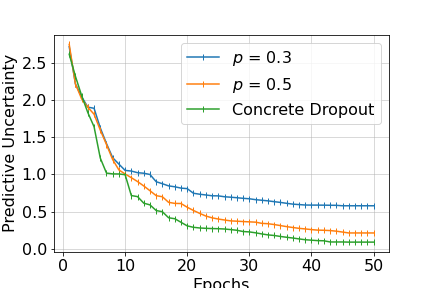}
    \caption{Plot showing uncertainty estimates for different values of dropout probability}
    \label{fig:unc_p}
\end{figure}

\textbf{Role of $\mathcal{L}_{\mathrm{MC}}(\theta)$:}  We see that on removing the Monte Carlo dropout loss term from the training loss $\mathcal{L}^{\text{DARTS}}_{val}$, the dropout probabilities in each of the layers are not updated, and hence the uncertainty is not calibrated. Therefore, we get a wrongful estimate of the uncertainty of the model. In Fig. \ref{fig:unc_p} we show this by repeating the same experiment with different values of dropout probabilities and show that each of the dropout probabilities gives rise to a different uncertainty estimate as discussed in \cite{gal2017concrete}.
We also note that if we remove the Monte Carlo loss function from the architecture search process while keeping the predictive variance term, we will get an architecture optimized for an uncalibrated noise. Hence, the optimal architecture found using this method may not be the best architecture to minimize the noise.

\begin{figure}
    \centering
    \includegraphics[width= 0.9\columnwidth]{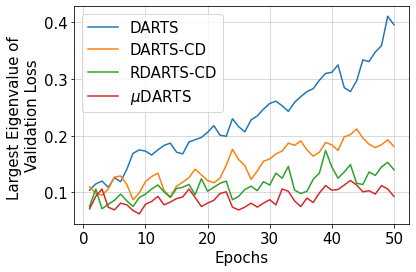}
    \caption{Plot showing largest eigenvalue of the Hessian of the validation loss function for increasing epochs for CIFAR 10 dataset}
    \label{fig:hess}
\end{figure}

\subsubsection{Robustness.}
We empirically evaluate the robustness of the architecture search methods by estimating the largest eigenvalue of the Hessian of the validation loss function ($\mathcal{L}_{\text{valid}}$) for each method.  Table ~\ref{tab:eigen_search} shows that the $\mu$DARTS method has a smaller value of the largest eigenvalue of the Hessian of the validation loss compared to the other methods making it more robust. We plot the evolution of the largest eigenvalue of the Hessian of the validation loss in Fig. \ref{fig:hess}. We see that while for the standard DARTS method, the largest eigenvalue keeps increasing for increasing epochs. However, the largest eigenvalue for the Hessian of the validation loss function of the $\mu$DARTS method increases very slowly and is much lower than the other methods discussed. The RDARTS-CD can generate similar robustness but requires the implementation of early stopping. The empirical analysis verifies that including predictive variance $\left(\operatorname{Var}_{p(\mathbf{y} | \mathbf{x})}^{model}(\alpha, w^{*}(\alpha)) \nonumber\right)$ in the validation loss $\left(\mathcal{L}_{\text{valid}}\right)$ improves robustness of the architecture search method (an analytical proof of this property is given in the Appendix assuming a linear model). 

\begin{table}
\centering
\caption{Robustness of Architecture Search Methods (Largest Eigenvalues of the $\nabla_{w}^{2} \mathcal{L}_{\text{valid}}$)}
\label{tab:eigen_search}
\resizebox{\columnwidth}{!}{%
\begin{tabular}{c||c c c||c}
\hline \hline
\textbf{Benchmark}        & \textbf{DARTS}&
\begin{tabular}[c]{@{}c@{}}\textbf{RDARTS-CD} \\ \end{tabular} &  \begin{tabular}[c]{@{}c@{}}\textbf{DARTS-CD} \\ \end{tabular} & 
\begin{tabular}[c]{@{}c@{}}$\mu$\textbf{DARTS} \\ \textbf{(This Work)}\end{tabular}\\ \hline \hline
\textbf{CIFAR10} & 0.396 & 0.110 & 0.142 & 0.097 \\ 
\textbf{CIFAR100} & 0.773 & 0.504 & 0.578 & 0.492 \\ 
\textbf{SVHN} & 0.202 & 0.045 & 0.079 & 0.021 \\ 
\textbf{ImageNet} & 0.202 & 0.045 & 0.079 & 0.021 \\\hline \hline
\end{tabular}%
}
\end{table}

\subsubsection{Convergence and Runtime Analysis}
\begin{figure}
    \centering
    \includegraphics[width=0.9\columnwidth]{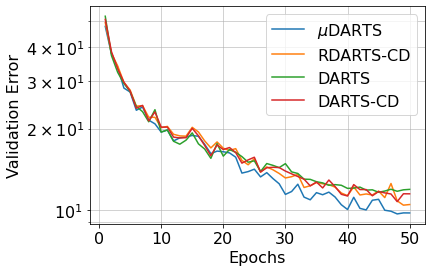}
    \caption{Convergence of Architecture Search methods for CIFAR10 dataset}
    \label{fig:val}
\end{figure}

In this section, we discuss the run times and the convergence of the architecture search method. Fig. \ref{fig:val} shows the validation error of the search models for each epoch for the CIFAR 10, CIFAR 100, SVHN, and the ImageNet datasets. $\mu$DARTS shows a faster convergence (or lower validation loss) than other architectures. Table \ref{tab:time} shows that the run-time of $\mu$DARTS is similar to DARTS, DARTS-CD, and RDARTS-CD.

\begin{table}[]

\caption{Run-time in GPU hours for different architecture search methods}
\centering

\resizebox{\columnwidth}{!}{%
\begin{tabular}{c|c c c c}
\hline \hline
 \textbf{Dataset}& \textbf{CIFAR10} & \textbf{CIFAR100} & \textbf{SVHN} & \textbf{ImageNet} \\ \hline\hline
\textbf{DARTS} & 9.3 & 37.5 & 5.1 & 102.3\\ 
\textbf{RDARTS-CD} & 9.7 & 37.6& 5.9 & 111.7\\ 
\textbf{DARTS-CD} & 9.4 & 37.5& 5.2 & 104.9\\ 
$\mu$\textbf{DARTS} & 9.6 & 37.7& 5.3 & 109.5\\ \hline \hline
\end{tabular}%
}
    \label{tab:time}
\end{table}

\begin{table*}
\centering
\caption{Performance of Final Architectures} 
\label{tab:eigen_arch}
\resizebox{0.65\textwidth}{!}{%
\begin{tabular}{c||c c c||c}
\hline \hline
\textbf{Benchmark}        & \textbf{DARTS} & 
\begin{tabular}[c]{@{}c@{}}\textbf{RDARTS-CD} \\ (Ours)\end{tabular} &  \begin{tabular}[c]{@{}c@{}}\textbf{DARTS-CD} \\ (Ours)\end{tabular} & \begin{tabular}[c]{@{}c@{}}$\mu$\textbf{DARTS} \\ \textbf{(This Work)}\end{tabular}\\ \hline \hline
\multicolumn{5}{c} {Largest Eigenvalues of the $\nabla_{w}^{2} \mathcal{L}_{\text{training}}$} \\ \hline 
\textbf{CIFAR10} & 0.217 & 0.093 & 0.102 & 0.089 \\ 
\textbf{CIFAR100} & 0.545 & 0.419 & 0.503 & 0.395 \\ 
\textbf{SVHN} & 0.176 & 0.039 & 0.061 & 0.017\\
\textbf{ImageNet} & 0.176 & 0.039 & 0.061 & 0.017 \\\hline \hline
\multicolumn{5}{c}{Testing Errors (Mean $\pm$ Standard Deviation)} \\ \hline
\textbf{CIFAR10}    &       5.68 $\pm$ 0.27      &   5.11 $\pm$ 0.21  &       5.71 $\pm$ 0.24 &   3.78 $\pm$ 0.20                                                      \\ 
\textbf{CIFAR100}        &      24.63  $\pm$  0.81    &    22.67 $\pm$ 0.66 & 25.19 $\pm$ 0.74 &   20.02 $\pm$  0.40   \\ 
\textbf{SVHN}        &      4.67  $\pm$  0.19    &    3.93  $\pm$ 0.15 & 4.48 $\pm$ 0.17 &   2.12  $\pm$ 0.127   \\ 
\textbf{ImageNet}        &   $29.37 \pm 5.03$       &   27.14   $\pm$ 3.97 &  28.71 $\pm$ 4.98 &   25.36  $\pm$ 3.92   \\ \hline \hline
\end{tabular}%
}
\end{table*}

\subsection{Comparative Analysis of the Final Architecture}
\textbf{Performance of Final Architectures.} 
 We compare the performance of the final architectures obtained from the architecture search methods by computing the largest eigenvalue of the Hessian of the training loss function $\mathcal{L}_{\text{train}}$. Table \ref{tab:eigen_arch} shows that the final architecture obtained from $\mu$DARTS has a lower eigenvalue of the training loss function indicating convergence to a flatter minima. We further evaluate robustness considering the mean and standard deviation of errors by training the final architecture multiple times and testing it on the test dataset. Table \ref{tab:eigen_arch} shows that $\mu$DARTS has a lower mean and standard deviation of the testing error compared to DARTS, DARTS-CD, and RDARTS-CD. Hence, we empirically verify that using the Monte-Carlo dropout loss as a regularizer in the training loss in $\mu$DARTS, instead of using an L2 regularizer as done in RobustDARTS ~\cite{arber2019understanding}, improves the robustness of the final architecture. 
These results prove that the $\mu$DARTS architecture search method converges to a flatter minima for each iteration of the bi-level optimization. 

\textbf{Importance of MC Dropout Loss within Bi-level Optimization.} Also, it is to be noted that in the DARTS-CD method, we are manually adding the Concrete Dropout layers after the architecture search process. So, the comparative analysis of the DARTS-CD and the $\mu$DARTS shows us that just adding the concrete dropout without solving the bi-level optimization problem does not provide us with the optimal solution.

\begin{table*}[]
\centering

\caption{Comparison of Accuracy and Uncertainty of Final Models from Different Architecture Search Processes (All models are trained for 50 epochs and batch size of 64)}
\label{tab:acc_unc_full}
\resizebox{\textwidth}{!}{%
\begin{tabular}{c||c||c c||c c||c c||c c}
\hline \hline
\textbf{Model} & \textbf{FLOPS}(M) & \multicolumn{2}{c||}{\textbf{CIFAR10}} & \multicolumn{2}{c||}{\textbf{CIFAR100}} & \multicolumn{2}{c||}{\textbf{SVHN}}& \multicolumn{2}{c}{\textbf{ImageNet}} \\ 
\multicolumn{1}{l||}{} & & \multicolumn{1}{l}{Accuracy} & \multicolumn{1}{l||}{Uncertainty} & \multicolumn{1}{l}{Accuracy} & \multicolumn{1}{l||}{Uncertainty} & \multicolumn{1}{l}{Accuracy} & Uncertainty & \multicolumn{1}{l}{Accuracy} & Uncertainty \\ \hline \hline
\textbf{MobileNetv2} \cite{sandler2018mobilenetv2} & 569 & 88.79 $\pm$ 0.38 & 0.915 & 68.54 $\pm 0.56$ & 1.266 & 93.67 $\pm 0.22$ & 0.247 & 68.83 $\pm $ 3.99 & 1.853     \\ 
\textbf{VGG16} \cite{simonyan2014very} & 160 & 84.11 $\pm$ 0.41 & 1.034 & 71.19 $\pm$ 0.52 & 1.482 & 93.15 $\pm$ 0.21 &  0.344 & 62.46 $\pm$ 2.11 & 1.897   \\ 
\textbf{ResNet20} \cite{he2016deep} & 320 & 86.47 $\pm$ 0.37 & 0.895 & 75.22 $\pm$ 0.46 & 1.246 & 93.74 $\pm$ 0.16 &  0.238  & 65.11 $\pm$ 2.02 & 1.816  \\ 
\textbf{EfficientNet-B0} \cite{tan2019efficientnet} & 390 & 89.01 $\pm$ 0.47 & 1.059 & 76.02 $\pm$ 0.55 & 1.215 & 94.87 $\pm$ 0.24 &  0.291   & 70.83 $\pm$ 4.15 & 1.893  \\ 
\hline \hline
\textbf{DARTS} \cite{liu2018darts} & 595 & 94.32 $\pm$ 0.27 & N/A & 75.37 $\pm$ 0.81 & N/A & 95.33 $\pm$ 0.19 & N/A  & 70.63 $\pm$ 5.03 & NA   \\ 

\textbf{P-DARTS-CD} & 532 & 95.32 $\pm$ 0.78 & 0.496 & 78.94 $\pm$ 0.89 & 0.883 & 97.79 $\pm$ 0.33 &  0.266   & 72.13 $\pm$ 4.07 & 1.854  \\ 
\textbf{PC-DARTS-CD} & 557 & 96.47 $\pm$ 0.81 & 0.463 & 80.01 $\pm$ 0.95 & 0.964 & 96.82 $\pm$ 0.36 &  0.279   & 72.81 $\pm$ 4.12 & 1.896  \\ 
\textbf{GOLD-NAS-A-CD} & 278 & 95.06 $\pm$ 1.03 & 0.412 & 78.97 $\pm$ 1.15 & 0.992 & 96.13 $\pm$ 0.52 & 0.333   & 69.56 $\pm$ 4.95 &  2.038 \\ 
\textbf{GOLD-NAS-I-CD} & 463 & 93.21 $\pm$ 0.91 & 0.519 & 79.15 $\pm$ 1.06 & 1.003 & 96.71 $\pm$ 0.47 &  0.303   & 72.21 $\pm$ 4.13 &  1.755 \\ 
\textbf{ASAP-CD} & 573 & 95.72 $\pm$ 1.15 & 0.481 & 79.27 $\pm$ 1.11 & 0.992 & 98.01 $\pm$ 0.63 &  0.484   & 73.96 $\pm$ 4.28  & 1.812  \\ 
\textbf{DARTS-CD}& 598 & 94.29 $\pm$ 0.24   & 0.474  &  74.81 $\pm$ 0.74  & 1.097  &  95.52 $\pm$ 0.17 &  0.195  & 71.29 $\pm$ 4.98 & 1.712  \\ 
\textbf{RDARTS-CD}& 605 & 94.89 $\pm$ 0.21 & 0.396 & 77.34 $\pm$ 0.66 & 0.815 & 96.07 $\pm$ 0.15 & 0.189  & 72.86 $\pm$ 3.97 & 1.684   \\ \hline \hline
\begin{tabular}[c]{@{}c@{}}$\mu$\textbf{DARTS} \\ (This Work)\end{tabular}& 602 & 96.22 $\pm$ 0.20 & 0.162 & 79.98 $\pm$ 0.40 & 0.561 & 97.88 $\pm$ 0.11 & 0.127  & 74.64 $\pm$ 3.92 & 1.523   \\ \hline \hline
\end{tabular}%
 }
\end{table*}

\subsection{Comparative Analysis of the Final Models} 

\subsubsection{Accuracy and Uncertainty}
We compare the performance of $\mu$DARTS considering CIFAR10 , CIFAR100 ~\cite{krizhevsky2009learning}, SVHN ~\cite{netzer2011reading} and ImageNet ~\cite{deng2009imagenet} datasets. All experiments are performed using PyTorch ~\cite{paszke2019pytorch} based models. 
For each dataset, we estimate the accuracy and uncertainty of the final architectures obtained from $\mu$DARTS with the following models:  
(1) architectures obtained from standard DARTS, (2) architectures obtained from the other baseline architecture search methods, and (3) existing deep networks for image classification with Concrete dropout. We calculate the model uncertainty by the number of Monte Carlo passes of the network and getting the predictive variance (see equation (\ref{eq:var})).

As examples of image classification network, we consider MobileNetv2 ~\cite{sandler2018mobilenetv2} , VGG16~\cite{simonyan2014very}, ResNet20 ~\cite{he2016deep}, EfficientNet ~\cite{tan2019efficientnet}. All networks were implemented from the source code. We used the standard models for each of them and implemented them in PyTorch. We also added a Concrete dropout-based ReLu layer instead of the ReLu layer in the original implementation of each of the models. An MC dropout loss is also added to the loss function to calculate the optimal dropout probabilities to get a tighter bound on the uncertainty estimate of the model. The modified models were re-trained, and uncertainty estimations were performed considering multiple MC samples.

We obtained the confidence interval of the performances of the architectures searched using the architecture search methods by calculating the mean and standard deviation of the model accuracy. We did this by re-training and re-evaluating the final searched model 10 times with different initializations. The observed mean and variance of the accuracy of the models and the mean of their uncertainty are shown in Table \ref{tab:acc_unc_full}. The table shows that the $\mu$DARTS outperforms the standard differentiable architecture search methods like DARTS-CD, RDARTS-CD, P-DARTS-CD, PC-DARTS-CD, GOLD-NAS-CD, and ASAP-CD. The architecture obtained with $\mu$DARTS also outperforms standard non-NAS-based architectures like MobileNetv2, VGG16, ResNet20, and EfficientNet-B0. It is to be noted here that all of the baseline models shown in Table V are equipped with concrete dropout to calculate the uncertainty of the model. We see that under these training conditions, the CIFAR-10 results of $\mu$DARTS are much higher (96.22\%) than the standard DARTS (94.32\%). A similar trend can be seen for ImageNet (74.64\% for $\mu$DARTS compared to 70.63\% for DARTS). This shows that the proposed $\mu$DARTS method can achieve good accuracy and uncertainty scores with minimal training (50 epochs) compared to the other baselines.
A reason for the better performance of the model on the ImageNet dataset in comparison to CIFAR10 might be attributed to the better regularization of the loss functions obtained due to the addition of the predictive variance and the Monte Carlo loss functions help in a more efficient search for architecture.
Also, adding the regularization terms helps stabilize the $\mu$DARTS search by implicitly regularizing the Hessian of the loss functions. We see that the implicit regularization leads to a smaller dominant eigenvalue of the Hessian, which serves as a proxy for the sharpness and thus leads to a more generalizable architecture.

\begin{table*}[t]
\centering
\caption{Accuracy and Uncertainty of Different DNN Models under Input Noise}
\label{tab:noisy1}
\resizebox{0.85\textwidth}{!}{%
\begin{tabular}{c||c c||c c||c c}
\hline \hline
\textbf{Model | SNR} & \multicolumn{2}{c||}{\textbf{40dB}} & \multicolumn{2}{c||}{\textbf{30dB}} & \multicolumn{2}{c}{\textbf{20dB}} \\ 
\multicolumn{1}{l||}{} & \multicolumn{1}{l}{Accuracy} & \multicolumn{1}{l||}{Uncertainty} & \multicolumn{1}{l}{Accuracy} & \multicolumn{1}{l||}{Uncertainty} & \multicolumn{1}{l}{Accuracy} & Uncertainty  \\ \hline \hline
\multicolumn{7}{c}{Accuracy and Uncertainty of Different Network Architectures under Input Noise for CIFAR 10} \\ \hline \hline
\textbf{DARTS} & 88.38 & N/A & 82.69 & N/A & 40.87 & NA   \\ 
\textbf{DARTS-CD} & 89.16 & 0.574 & 82.17 & 0.913 & 44.11 & 1.502   \\ 
\textbf{RDARTS-CD} & 90.88   & 0.488 &  82.75  & 0.899  &  44.29 &  1.479 \\ \hline \hline
\begin{tabular}[c]{@{}c@{}}$\mu$\textbf{DARTS}\end{tabular} & 92.04 & 0.198 & 84.16 & 0.397 & 49.34 & 1.206   \\ \hline \hline

\multicolumn{7}{c}{Accuracy and Uncertainty of Different Network Architectures under Input Noise for CIFAR 100} \\ \hline \hline
\textbf{DARTS} & 76.21 & N/A & 71.37 & N/A & 37.17 & NA   \\ 
\textbf{DARTS-CD} & 76.89 & 0.632 & 71.04 & 1.174 & 38.41 & 2.047   \\ 
\textbf{RDARTS-CD} & 77.39   & 0.599 &  72.58  & 1.007  &  39.16 &  1.992 \\ \hline \hline
\begin{tabular}[c]{@{}c@{}}$\mu$\textbf{DARTS}\end{tabular} & 78.54 & 0.483 & 75.82 & 0.397 & 46.22 & 1.632   \\ \hline \hline
\multicolumn{7}{c}{Accuracy and Uncertainty of Different Network Architectures under Input Noise for SVHN} \\ \hline \hline
\textbf{DARTS} & 94.77 & N/A & 89.19 & N/A & 70.22 & NA   \\ 
\textbf{DARTS-CD} & 94.98 & 0.148 & 90.33 & 0.396 & 71.68 & 0.835   \\ 
\textbf{RDARTS-CD} & 95.03   & 0.106 &  91.00  & 0.313  &  72.15 &  0.794 \\ \hline \hline
\begin{tabular}[c]{@{}c@{}}$\mu$\textbf{DARTS}\end{tabular} & 97.34 & 0.095 & 92.3 & 0.199 & 75.45 &  0.552  \\ \hline \hline
\end{tabular}%
 }
\end{table*}

\begin{table*}[h]
\centering
\caption{Performance  of  Final  DNN under Parameter noise}
\label{tab:param_noise}
\resizebox{0.75\textwidth}{!}{%
\begin{tabular}{c||c c || c c|| c c}
\hline\hline
                   & \multicolumn{2}{c|}{\textbf{CIFAR 10}}                                               & \multicolumn{2}{c|}{\textbf{CIFAR 100}}                                                & \multicolumn{2}{c}{\textbf{SVHN}}                                                    \\ \hline \hline
        & \textbf{Accuracy} & \textbf{Uncertainty} & \textbf{Accuracy}   & \textbf{Uncertainty} & \textbf{Accuracy}  & \textbf{Uncertainty} \\ \hline\hline
\textbf{DARTS}     & 90.17 $\pm$ 0.298  &   N/A                                                               & 74.13  $\pm$  0.785 &    N/A                                                              & 93.19  $\pm$  0.211 &       N/A                                                           \\ 
\textbf{DARTS-CD}  & 91.86 $\pm$ 0.297  &   0.312                                                               & 74.68 $\pm$ 0.587   &     0.638                                                             & 93.91  $\pm$ 0.206  &      0.201                                                            \\ 
\textbf{RDARTS-CD} & 92.78 $\pm$ 0.29   &     0.217                                                             & 75.76 $\pm$ 0.591   & 0.411                                                                 & 94.11 $\pm$ 0.210     &       0.109                                                           \\ \hline\hline
$\mu$\textbf{DARTS}   & 94.12 $\pm$ 0.25   &   0.134                                                               & 79.89 $\pm$  0.491  &      0.398                                                            & 95.14  $\pm$ 0.17   &      0.074                                                            \\ \hline\hline
\end{tabular}%
}
\end{table*}

\subsubsection{Input Noise Tolerance of Final DNN Models}
We study the accuracy and uncertainty of the final DNN models obtained from various architecture search methods under Gaussian noise to the input images. We used the CIFAR10, CIFAR100, and SVHN datasets and compared varying Signal-to-Noise Ratio (SNR). We do not re-train the models with noisy images. Instead, noisy images are only applied during inference [Table \ref{tab:noisy1}]. The added predictive variance term and the Monte-Carlo loss terms act as a regularizer, improving the performance compared to the standard methods. We observe that the $\mu$DARTS process has higher accuracy and lower uncertainty under noise than DNN models from other architecture search methods.

\subsubsection{Tolerance of Final DNN Models to Noisy Parameters}

We test the architecture's performance when the model parameters are perturbed by a small gaussian noise $\sim\mathcal{N}(0,1)$. We hypothesize that a more stable neural architecture will lower the testing error and uncertainty variance.
Table \ref{tab:param_noise} shows that the DNN model generated by the $\mu$DARTS method gives the least variance. Hence, we conclude that the $\mu$DARTS final architecture is stable and can handle noise perturbations very well. Comparing the results of Tables \ref{tab:acc_unc_full} and \ref{tab:param_noise}, we see that with the inclusion of the parameter noise, though the accuracy for all the models fall and the uncertainty increases, the change in the accuracy and uncertainty for the $\mu$DARTS model [e.g. CIFAR10: $\Delta \text{Mean Accuracy} = 2.61, \Delta \text{Variance}=0.045$] is lower than the change in other networks. Hence, we can conclude that the $\mu$DARTS model is more resilient to parameter noise.

\section{Conclusions}
In this paper, we proposed a novel method, referred to as $\mu$DARTS, to search for a neural architecture that simultaneously improves the accuracy and the uncertainty of the final neural network architecture. $\mu$DARTS uses concrete dropout layers within the cells of a neural architecture search framework, and adds predictive variance and Monte-Carlo dropout losses as regularizers in the validation and training losses respectively. We experimentally demonstrate that $\mu$DARTS improves the performance of the neural architecture search and the final architecture found using the architecture search by showing that the optimization problem converges to a flat minima. We also empirically show that the final architecture is stable when perturbed with input and parameter noise.


\section*{Appendix}

\subsection{Implicit Regularization of the Hessian} 

The addition of the regularizer to the loss function of the $\mu$DARTS architecture search method induces implicit regularization on the Hessian matrix. It has been empirically pointed out that the dominant eigenvalue of $\nabla_{A}^{2} L_{\mathrm{val}}(w, A)$ (spectral norm of Hessian) is highly correlated with the generalization quality of DARTS solutions \cite{arber2019understanding}. In standard DARTS training, the Hessian norm grows exponentially leading to bad test performance. Chen et al. \cite{chen2020stabilizing} plot this Hessian norm during the training procedure and found that DARTS with perturbation-based regularization consistently reduce the Hessian norms during the training procedure. They also showed the spectral norm of Hessian is correlated with the solution quality, and that perturbation-based regularization of the DARTS search strategy can implicitly control the Hessian norm.

The update of $w$ in $\mu$DARTS can thus implicitly control the trace norm of the Hessian of the loss function. If the matrix is close to positive semi-definite, this is approximately regularizing the (positive) eigenvalues of $\nabla_{\alpha}^{2} L^{\mu\text{DARTS}}_{\mathrm{val}}(w, \alpha)$. Therefore, $\mu$DARTS reduces the Hessian norm through its training procedure.

\subsection{Comparison of Stability using Largest Eigenvalues}
In \cite{arber2019understanding}, the authors empirically showed the instability of the DARTS method is related to the norm of Hessian $\nabla_{\boldsymbol{\alpha}}^2\mathcal{L}_{\text{valid}}$. Chen et al. \cite{chen2020stabilizing} verified this by plotting the validation accuracy landscape of the DARTS and showed it is extremely sharp and thus, even a small perturbation in $\boldsymbol{\alpha}$ can drastically change the validation accuracy.

Here we prove that the eigenvalue of the Hessian of validation loss in $\mu$DARTS ($\nabla_{\boldsymbol{\alpha}}^{2} \mathcal{L}_{\text{valid}}^{\text{$\mu$DARTS}}$) is lower than the eigenvalue of the Hessian of validation loss in DARTS ($\nabla_{\boldsymbol{\alpha}}^{2} \mathcal{L}_{\text{valid}}^{\text{DARTS}}$). 
\begin{lemma}
The largest eigenvalue of the Hessian of the validation loss of DARTS is given by, \begin{align} \lambda_{\text{max}}(\nabla_{\alpha}^{2}\mathcal{L}_{\text{valid}}^{\text{DARTS}}) = \frac{\max \sigma_i}{N}\lambda_{\max} (\sum_{i=1}^{N} \mathbf{x}_{i} \mathbf{x}_{i}^{T}), \nonumber \\
\text{where, } \sigma_{i}=\sigma\left(\mathbf{x}_{i}^{T} \boldsymbol{\alpha}\right)\left(1-\sigma\left(\mathbf{x}_{i}^{T} \boldsymbol{\alpha}\right)\right)\end{align} 
\label{lemma:DARTS}
\end{lemma}

\textbf{Proof.} We consider $\mathbf{x}_{i}$ is the input vector,  $\alpha$ is the weight matrix of architectural parameter, and $\mathbf{y}_i$ is the vector denoting the labels of the classes.
For simplicity, we consider a linear model $\mathbf{x}_{i}^{T} \boldsymbol{\alpha}$ for this proof. 
Considering the cross entropy loss as the validation loss ($\mathcal{L}_{\text{valid}}$) we obtain:
\begin{align}\mathcal{L}_{\text{valid}}^{\text{DARTS}}=&-\frac{1}{N} \sum_{i=1}^{N} \mathbf{y}_{i} \log \left(\sigma\left(\mathbf{x}_{i}^{T} \boldsymbol{\alpha}\right)\right)+ \nonumber \\
&\left(1-\mathbf{y}_{i}\right) \log \left(1-\sigma\left(\mathbf{x}_{i}^{T} \boldsymbol{\alpha}\right)\right) 
\end{align}
\begin{align}
\nabla_{\alpha}^{2} \mathcal{L}_{\text{valid}}^{\text{DARTS}}&=\frac{1}{N} \sum_{i=1}^{N} \sigma\left(\mathbf{x}_{i}^{T} \boldsymbol{\alpha}\right)\left(1-\sigma\left(\mathbf{x}_{i}^{T} \boldsymbol{\alpha}\right)\right) \mathbf{x}_{i} \mathbf{x}_{i}^{T}\nonumber \\
&=\frac{1}{N} \sum_{i=1}^{N} \sigma_{d}\mathbf{x}_{i} \mathbf{x}_{i}^{T}
\end{align}
where, $\sigma(.)$ is the sigmoid function, $\sigma_{d}=\sigma\left(\mathbf{x}_{i}^{T} \boldsymbol{\alpha}\right)\left(1-\sigma\left(\mathbf{x}_{i}^{T} \boldsymbol{\alpha}\right)\right)$ for each $i$. 
We know the Rayleigh quotient of $\nabla_{\alpha}^{2} \mathcal{L}_{\text{valid}}^{\text{DARTS}}$ is given by: 
\begin{align}
\mathcal{R}_{\text{DARTS}}(\mathbf{z})&=\mathbf{z}^{T} \nabla_{\alpha}^{2} \mathcal{L}_{\text{valid}}^{\text{DARTS}} \mathbf{z} \nonumber \\
&=\mathbf{z}^{T}\left(\frac{1}{N} \sum_{i=1}^{N} \sigma_{d} \mathbf{x}_{i} \mathbf{x}_{i}^{T}\right) \mathbf{z} \nonumber \\
&=\frac{1}{N}\left(\mathbf{z}^{T}\left(\sum_{i=1}^{N} \sigma_{d}\mathbf{x}_{i} \mathbf{x}_{i}^{T}\right) \mathbf{z}\right)
\label{eq:r_loss}
\end{align}

where, $\mathbf{z}$ is any unit-length vector. Assuming maximum of $\sigma_{d}$ is represented as $\sigma_{d}^{max}$, we obtain:
\begin{align}
\mathcal{R}_{\text{DARTS}}(\mathbf{z}) &\le \frac{\sigma_{d}^{max}}{N}\left(\mathbf{z}^{T}\left(\sum_{i=1}^{N} \mathbf{x}_{i} \mathbf{x}_{i}^{T}\right) \mathbf{z}\right)  \nonumber 
= \frac{\sigma_{d}^{max}}{N}\mathcal{R}_{M_{\mathbf{x}}}(\mathbf{z})
\end{align}

where, $\mathcal{R}_{\text{M}}(\mathbf{z})$ is the Raleigh quotient of the matrix $M_{\mathbf{x}}=\sum_{i=1}^{N} \mathbf{x}_{i} \mathbf{x}_{i}^{T}$ as given by: 
\begin{align}
\mathcal{R}_{\text{M}}(\mathbf{z})=\mathbf{z}^{T}\left(\sum_{i=1}^{N} \mathbf{x}_{i} \mathbf{x}_{i}^{T}\right) \mathbf{z} \nonumber &=\sum_{i=1}^{N} \left(\mathbf{z}^{T} \mathbf{x}_{i}\right)\left(\mathbf{x}_{i}^{T} \mathbf{z}\right) \nonumber \\
&= \sum_{i=1}^{N} \left(\mathbf{z}^{T} \mathbf{x}_{i}\right)^{2} \ge 0
\label{eq:r_sum}
\end{align}



Since, the maximum eigenvalue of a symmetric matrix ($A$) is equal to the maximum value of its Raleigh quotient $\left(\lambda_{\max}(A)=\mathcal{R}_{A}^{max})\right)$, we obtain: \begin{align}
\lambda_{\text{max}}(\nabla_{\alpha}^{2}\mathcal{L}_{\text{valid}}^{\text{DARTS}}) &= \mathcal{R}_{\text{DARTS}}^{\max}=\frac{ \sigma_{d}^{\max}}{N}\mathcal{R}_{M
}^{\max} \nonumber \\
&=\frac{ \sigma_{d}^{\max}}{N} \lambda_{\max} \left(\sum_{i=1}^{N} \mathbf{x}_{i} \mathbf{x}_{i}^{T}\right)
\label{eq:spectral_DARTS}
\end{align}

\textbf{Corollary.}  The validation loss function of the standard DARTS ($\mathcal{L}_{\text{valid}}^{\text{DARTS}}$) method is convex in nature.

\textbf{Proof.} We know that, the smallest eigenvalue of any square symmetric matrix is given as the minimum of the Rayleigh quotient of that matrix. Since $\sigma_{d} \geq 0$, from equations \ref{eq:r_loss} and \ref{eq:r_sum}, we obtain $\mathcal{R}_{\text{DARTS}}(\mathbf{z}) \geq 0$. Hence, the smallest possible eigenvalue of $\nabla_{\alpha}^{2} \mathcal{L}_{\text{valid}}^{\text{DARTS}}$ must be zero or positive which implies that the validation loss function is convex.

\begin{lemma}
The largest eigenvalue of the Hessian of the validation loss of $\mu$DARTS is given by, \[ \lambda_{\text{max}}(\nabla_{\alpha}^{2}\mathcal{L}_{\text{valid}}^{\text{$\mu$DARTS}}) \le \frac{ \sigma_{ud}^{max}}{N}\lambda_{\text{max}}\left(\sum_{i=1}^{N} \mathbf{x}_{i} \mathbf{x}_{i}^{T}\right) \] 

\begin{align} 
 \text{ where, } \sigma_{ud} &= \sqrt{\sigma_{x^2}}\left(1-\sqrt{\sigma_{x^2}}\right)  +  4\alpha^2 \sigma_{x^2}(1- \sigma_{x^2})  \nonumber \\
&-8\alpha^2(\sigma_{x^2})^{2}(1- \sigma_{x^2}) 
 + 2\sigma_{x^2}(1 - \sigma_{x^2})   \nonumber \\
& -2(\sqrt{\sigma_{x^2}}(1-\sqrt{\sigma_{x^2}})+ \sqrt{\sigma_{x^2}}) 
  \nonumber \\
\sigma_{x^2} &= \sigma(\boldsymbol{\alpha} \mathbf{x}_{i} \mathbf{x}_{i}^{T} \boldsymbol{\alpha}^{T}) \text{ and } \sigma_{x} = \sigma(\mathbf{x}_{i}^{T} \boldsymbol{\alpha})  \nonumber
\end{align}

\label{lemma:muDARTS}
\end{lemma}

\textbf{Proof.} To estimate $\nabla_{\alpha}^{2} \mathcal{L}_{\text{valid}}^{\text{$\mu$DARTS}}$, we add the predictive variance term to the DARTS validation loss. 
\begin{align}
\operatorname{Var}_{p(\mathbf{y} | \mathbf{x})}^{model}(\boldsymbol{\alpha}) &= \mathbb{E}((\mathbf{x}_{i}^{T}\alpha)^2)  - \mathbb{E}(\mathbf{x}^{T}_{i}\alpha)^2
\nonumber \\
&= \frac{1}{T} \sum_{i=1}^{T}\sigma( \boldsymbol{\alpha} \mathbf{x}_{i} \mathbf{x}_{i}^{T} \boldsymbol{\alpha}^{T}) - (\frac{1}{T} \sum_{i=1}^{T}\sigma(\mathbf{x}_{i}^{T} \boldsymbol{\alpha}))^{2} \\
\nabla_{\alpha}^{2} \mathcal{L}_{\text{valid}}^{\text{$\mu$DARTS}} &= \nabla_{\alpha}^{2} \mathcal{L}_{\text{valid}}^{\text{DARTS}} +  \nabla_{\alpha}^{2} \operatorname{Var}_{p(\mathbf{y} | \mathbf{x})}^{model}(\boldsymbol{\alpha}) \nonumber\\
 &= \frac{1}{N} \sum_{i=1}^{N} \sigma_{x}\left(1-\sigma_{x}\right) \mathbf{x}_{i} \mathbf{x}_{i}^{T} \nonumber \\
+ \frac{1}{T} \sum_{i=1}^{T} \{ &[\sigma_{x^2}(1- \sigma_{x^2}) -2\sigma_{x^2}^{2}(1- \sigma_{x^2})]4\alpha^{T}\alpha \mathbf{x}_{i}\mathbf{x}^{T}_{i} \nonumber \\
& \quad \quad \quad \quad  \quad \quad  + \sigma_{x^2}(1 - \sigma_{x^2}) 2\mathbf{x}_{i}\mathbf{x}^{T}_{i} \} \nonumber\\ &- \frac{2}{T^{2}} [\sum_{i=1}^{T} \sigma_{x}(1- \sigma_{x})\sum_{i=1}^{T}\sigma_{x}(1- \sigma_{x})]\mathbf{x}_{i} \mathbf{x}_{i}^{T} \nonumber \\
-\frac{2}{T^{2}}[\sum_{i=1}^{T} \sigma_{x}& \sum_{i=1}^{T} \sigma_{x}(1- \sigma_{x}) -2(\sigma_{x})^2(1-\sigma_{x})] \mathbf{x}_{i} \mathbf{x}_{i}^{T}
\end{align}
Without loss of generality, we consider the case when $T=N$. Therefore, the Raleigh quotients of $\nabla_{\alpha}^{2}\mathcal{L}_{\text{valid}}^{\text{$\mu$DARTS}}$ can be computed as: 

\begin{align}
&\mathcal{R}_{\text{$\mu$DARTS}}(\mathbf{z})=\mathbf{z}^{T}\nabla_{\alpha}^{2} \mathcal{L}_{\text{valid}}^{\text{$\mu$DARTS}} \mathbf{z} \nonumber \\
 &= \mathbf{z}^{T}\Bigg( \frac{1}{N}\sum_{i=1}^{N}
 \bigg[
 \sigma_{x}\left(1-\sigma_{x}\right)  +  \Big(\sigma_{x^2}(1- \sigma_{x^2}) \nonumber \\
& \quad \quad \quad -2\sigma_{x^2}^{2}(1- \sigma_{x^2})\Big)4\alpha^{T}\alpha  + 2\sigma_{x^2}(1 - \sigma_{x^2}) \bigg] \mathbf{x}_{i}\mathbf{x}_{i}^{T} 
 \nonumber \\ 
  &- \frac{2}{N^2} \bigg[\sum_{i=1}^{N}  \sigma_{x}(1- \sigma_{x})\sum_{i=1}^{N}\sigma_{x}(1- \sigma_{x}) \nonumber \\
& + \sum_{i=1}^{N}  \sigma_{x} \sum_{i=1}^{N} \Big(\sigma_{x}(1- \sigma_{x}) -2(\sigma_{x})^2(1-\sigma_{x})\Big)\bigg] \mathbf{x}_{i}\mathbf{x}_{i}^{T}\Bigg)\mathbf{z} \nonumber 
\end{align}


Since $N$ is large, we note that $\sum_{i=1}^{N} \sigma_{x}(1- \sigma_{x}) \ge 1 ; \quad  \sum_{i=1}^{N}\big[ \sigma_{x}(1- \sigma_{x}) -2(\sigma_{x})^2(1-\sigma_{x})\big] \ge 1$. Thus, we can say that 
\begin{align}
(i) \sum_{i=1}^{N} \sigma_{x}(1- \sigma_{x}) \sum_{i=1}^{N} \sigma_{x}(1- \sigma_{x}) \ge \sum_{i=1}^{N} \sigma_{x}(1- \sigma_{x})  \\
(ii) \sum_{i=1}^{N} \sigma_{x} \sum_{i=1}^{N}\big[ \sigma_{x}(1- \sigma_{x}) -2(\sigma_{x})^2(1-\sigma_{x})\big]  \ge \sum_{i=1}^{N}\sigma_{x} 
\end{align}

\begin{table*}[]
\centering
\caption*{TABLE C.1: Comparison of Reported Test Errors of Models for Different Architecture Search Processes (The test errors are the values reported in the original papers)}
\label{tab:acc_comp}
\resizebox{0.9\textwidth}{!}{%
\begin{tabular}{c||c c c c || c c c c}
\hline \hline
\textbf{Model}  & \multicolumn{4}{c||}{\textbf{CIFAR 10}} & \multicolumn{4}{c}{\textbf{ImageNet}} \\

& \textbf{Params}(M) &  \textbf{Accuracy} & \textbf{Epochs} & \textbf{Batch Size} & \textbf{Params}(M) &  \textbf{Accuracy}  & \textbf{Epochs} & \textbf{Batch Size}\\ \hline \hline 
\textbf{P-DARTS} \cite{chen2019progressive} & 10.5 & 2.25 & 250 & 1024 & 5.4 & 24.1 & 250 & 1024 \\ 
\textbf{PC-DARTS} \cite{xu2019pc} & 3.6 & 2.57 & 600 & 128 & 5.3 & 24.2 & 250 & 1024   \\ 
\textbf{SNAS} \cite{xie2018snas} & 2.8 & 2.85 & 600 & 96 & 4.3 & 27.3 & 250 & 96    \\ 
\textbf{GOLD-NAS-A} \cite{bi2020gold} & 1.58 & 2.93 & 200 & 96 &-  & - & -  \\ 
\textbf{GOLD-NAS-Z} \cite{bi2020gold} & - & - & - & - & 6.4& 23.9  & 250 & 1024  \\ 
\textbf{ASAP} \cite{noy2020asap} & 6.0 & 1.68 & 1500 & 128& 6.0 & 24.4 & 250 & 64  \\ 
\textbf{DARTS} \cite{liu2018darts} & 3.3 & 2.76 & 600 & 96 & 4.7 & 26.7 & 250 & 128 \\ 
\textbf{RDARTS} \cite{arber2019understanding}& 3.1 & 2.97 & 50 & 64 & - & - & - \\ 
\hline \hline
\begin{tabular}[c]{@{}c@{}}$\mu$\textbf{DARTS}\\ (This Work)\end{tabular}& 4.1 & 3.78 & 50 & 32 & 5.3  & 25.36 & 50 & 64   \\ \hline \hline
\end{tabular}%
 }
\end{table*}

We take $\mathcal{R}_{\text{$\mu$DARTS}}(\mathbf{z})= \mathbf{z}^{T}\Bigg( \frac{1}{N}\sum_{i=1}^{N}
 \bigg[\sigma_{j}\bigg]\mathbf{x}_{i}\mathbf{x}_{i}^{T}\Bigg)\mathbf{z},$ where, $\sigma_{j}$ is the polynomial in $\sigma_x$, $\sigma_{x^2}$, and $\alpha$. Also, since $\sigma_{x}$ is convex, using Jensen's inequality ($\sigma_{x}^{2} \le \sigma_{x^2}$) we get:
\begin{align}
\sigma_{j} &\le \Big[\sqrt{\sigma_{x^2}}\left(1-\sqrt{\sigma_{x^2}}\right)  + 4\alpha^T \alpha \sigma_{x^2}(1- \sigma_{x^2}) \nonumber \\
&-8\alpha^T \alpha(\sigma_{x^2})^{2}(1- \sigma_{x^2}) + 2\sigma_{x^2}(1 - \sigma_{x^2})  
 \nonumber \\
&-2\big(\sqrt{\sigma_{x^2}}(1-\sqrt{\sigma_{x^2}})+ \sqrt{\sigma_{x^2}}\big)\Big]
\end{align}
Assuming, maximum value of $\sigma_{ud}$ and $\sigma_{j}$ are given by $\sigma_{ud}^{\max}$ and $\sigma_{j}^{\max}$, respectively, we obtain:
\begin{align}
 \mathcal{R}_{\text{$\mu$DARTS}}(\mathbf{z}) 
 &\le \frac{\sigma_{j}^{\max}}{N}\Bigg(\mathbf{z}^{T}\Bigg(\sum_{i=1}^{N}
\mathbf{x}_{i}\mathbf{x}_{i}^{T}\Bigg)\mathbf{z}\Bigg)
  \nonumber \\
&\le \frac{\sigma_{ud}^{\max}}{N}\Bigg(\mathbf{z}^{T}\Bigg(\sum_{i=1}^{N}
\mathbf{x}_{i}\mathbf{x}_{i}^{T}\Bigg)\mathbf{z}\Bigg)
 \end{align}
Using the relation between maximum eigenvalue and Ralyeigh quotient, similar to the Lemma \ref{lemma:DARTS}, the largest eigenvalue of the $\mu$DARTS method is given as:
\begin{align}
\lambda_{\text{max}}(\nabla_{\alpha}^{2}\mathcal{L}_{\text{valid}}^{\text{DARTS}}) &= \mathcal{R}_{\text{$\mu$DARTS}}^{\max}  \nonumber \\
&\le \frac{ \sigma_{ud}^{\max}}{N} \lambda_{\max} \left(\sum_{i=1}^{N} \mathbf{x}_{i} \mathbf{x}_{i}^{T}\right)
\label{eq:spectral_muDARTS}
\end{align}


\begin{lemma}
If $\lambda_{\text{max}}(\nabla_{\alpha}^{2}\mathcal{L}_{\text{valid}}^{\text{DARTS}})$ and $\lambda_{\text{max}}(\nabla_{\alpha}^{2}\mathcal{L}_{\text{valid}}^{\text{$\mu$DARTS}})$ be the maximum eigenvalues of the Hessian of the validation loss for \text{DARTS} and the \text{$\mu$DARTS}, respectively, then 

\begin{align}
\lambda_{\text{max}}(\nabla_{\alpha}^{2}\mathcal{L}_{\text{valid}}^{\text{$\mu$DARTS}}) &\leq \lambda_{\text{max}}(\nabla_{\alpha}^{2}\mathcal{L}_{\text{valid}}^{\text{DARTS}}) \nonumber
\end{align}
\end{lemma}



\textbf{Proof.} We represent $\sigma_{d}$ as: 
$\sigma_{d} = p(1-p)$, where $p=\sigma\left(\mathbf{x}_{i}^{T} \boldsymbol{\alpha}\right) \in [0,1]$ due to the properties of the sigmoid function. Moreover, since $\alpha$ is the weights in the neural network we have $\alpha < 1 \implies \alpha^T \alpha < 1$. Hence, by maximizing $\sigma_i$ with respect to $p$ we obtain:$\sigma_{d}^{\max} = 0.25$

We now represent $\sigma_{ud} = \sqrt{q}(1-\sqrt{q})+4\alpha^T \alpha[ q(1-q)- 2q^{2}(1-q)]+2 q(1-q)-2(\sqrt{q}(1-\sqrt{q})+ \sqrt{q} \}$, where $q=\sigma_{{x}^2} = \sigma(\boldsymbol{\alpha} \mathbf{x}_{i} \mathbf{x}_{i}^{T} \boldsymbol{\alpha}^{T}) \in [0,1] $. We now maximize $\sigma_j$ with respect to $q$. Finally, noting $\alpha^T \alpha  < 1$, we observe that $\sigma_{ud} \le 0$. Thus the function becomes: $\sigma_{ud}^{\max} < 0 < 0.25 = \sigma_{d}^{max} $. 
From equation (\ref{eq:r_sum}) we note that $\mathcal{R}_{\text{M}}^{max} \geq 0$. Therefore, using equations  (\ref{eq:spectral_DARTS}), (\ref{eq:spectral_muDARTS}), and the bound on $\sigma_{ud}^{\max}$, we get
\begin{equation}
\lambda_{\text{max}}(\nabla_{\alpha}^{2}\mathcal{L}_{\text{valid}}^{\text{$\mu$DARTS}})\leq \lambda_{\text{max}}(\nabla_{\alpha}^{2}\mathcal{L}_{\text{valid}}^{\text{DARTS}})
\end{equation}

\subsection{Comparison of Accuracy of Unmodified Models}

We demonstrate the comparative performance of the $\mu$DARTS method with respect to models searched using other architecture search methods with the training condition for each model as reported in the original paper.  Note that the original NAS and DARTS papers only reported accuracy of the final models (not uncertainty). The results are shown in Table C.1.  We see that the $\mu$DARTS, trained with only 50 epochs gives comparable performance to other networks trained for a much higher number of epochs and with a much higher batch size. This result further showcases the robustness and easy trainability of $\mu$DARTS.

\begin{IEEEbiography}[{\includegraphics[width=1in,height=1.25in,clip,keepaspectratio]{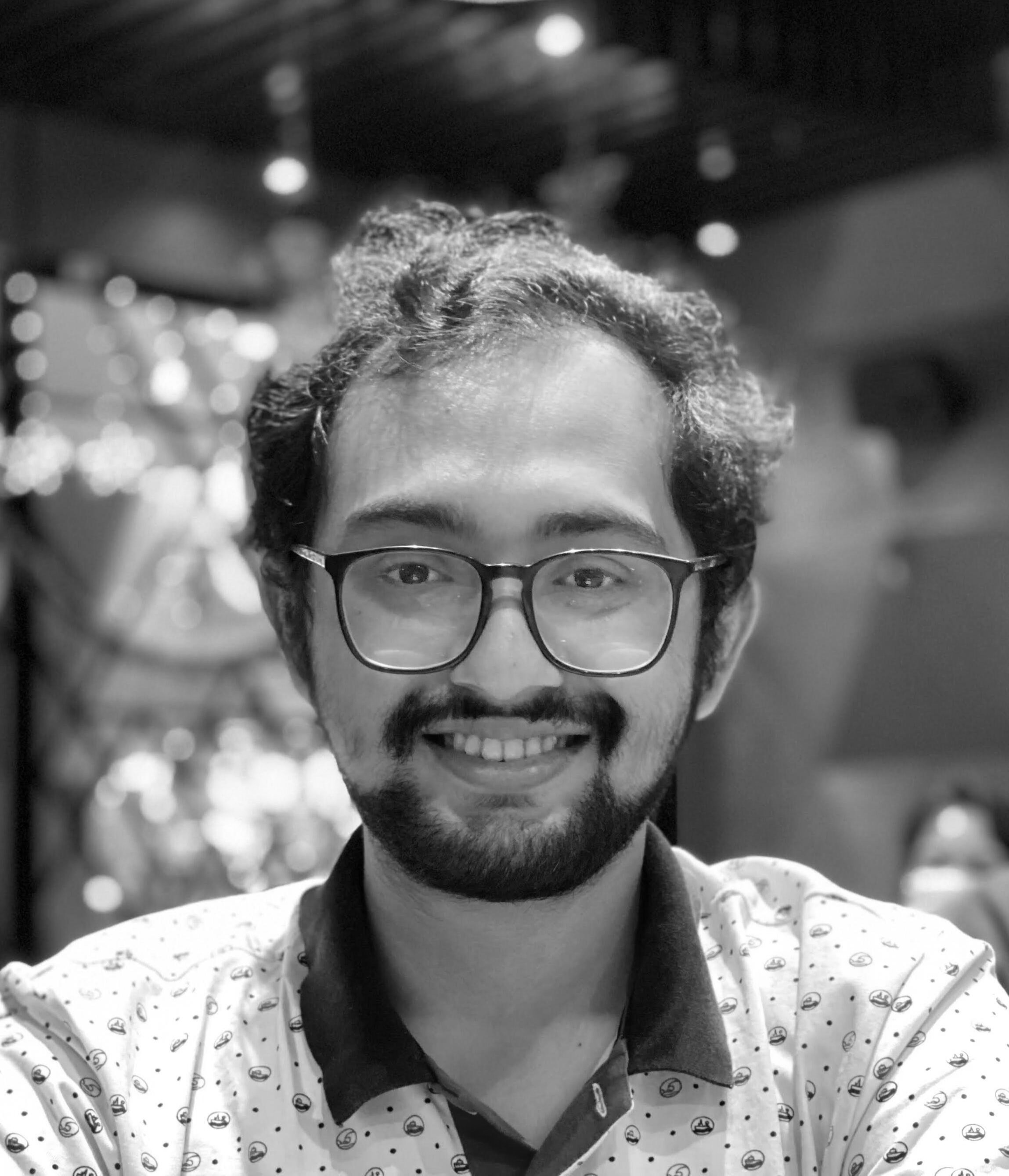}}]{Biswadeep Chakraborty} (Graduate Student Member, IEEE) received the B.E. degree in Electronics and Telecommunication engineering from Jadavpur University, Kolkata, India, in 2019. He is currently pursuing the Ph.D. degree in Electrical and Computer Engineering with the Georgia Institute of Technology, Atlanta, GA, USA, under the supervision of Prof. S. Mukhopadhyay. Before starting his Ph.D., he worked as a Research Assistant at the Singtel-National University of Singapore Lab. His current research interests lies in unsupervised learning methods using spiking neural networks for spatio-temporal prediction and control.
\end{IEEEbiography}

\begin{IEEEbiography}[{\includegraphics[width=1in,height=1.25in,clip,keepaspectratio]{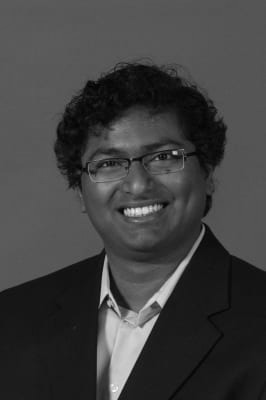}}]{Saibal Mukhopadhyay} (Fellow, IEEE) received the B.E. degree in Electronics and Telecommunication engineering from Jadavpur University, Kolkata, India, in 2000, and the Ph.D. degree in electrical and computer engineering from Purdue University, West Lafayette, IN, USA, in 2006. He was a Research Staff Member with the IBM Thomas J. Watson Research Center, Yorktown Heights, NY, USA, from August 2007 to September 2007. He is currently a Joseph M.
Pettit Professor with the School of Electrical and Computer Engineering, Georgia Institute of Technology, Atlanta, GA, USA. He has authored or coauthored over 200 articles in refereed journals and conferences, and holds five U.S. patents. His research interests include design of energy-efficient, intelligent, and secure systems in nanometer technologies. He was a recipient of the Office of Naval Research Young Investigator Award, in 2012, the National Science Foundation CAREER Award, in 2011, the IBM Faculty Partnership Award, in 2009 and 2010, the SRC Inventor Recognition Award in 2008, the SRC Technical Excellence Award, in 2005, and the IBM Ph.D. Fellowship Award, for years 2004 to 2005.
\end{IEEEbiography}

\EOD
\end{document}